\documentclass[letterpaper, 10 pt, conference]{ieeeconf}  
\IEEEoverridecommandlockouts                              
\usepackage{amsmath,amsfonts}
\usepackage{algorithmic}
\usepackage{algorithm}
\usepackage{array}
\usepackage{hyperref}
\usepackage{amsmath,empheq}
\usepackage{textcomp}
\usepackage{stfloats}
\usepackage{balance}
\usepackage{url}
\usepackage{enumerate}
\usepackage{wrapfig}
\usepackage{verbatim}
\usepackage{amsmath}
\usepackage{graphicx}
\usepackage{svg}
\usepackage{cite}
\usepackage{hyperref}

\hypersetup{
  colorlinks=true,
  linkcolor=black,
  urlcolor=black,
}
\usepackage{color}
\begin{document}


\title{A Multipurpose Interface for Close- and Far-Proximity\\Control of Mobile Collaborative Robots}

\author{Hamidreza Raei$^{1, 2}$, Juan M. Gandarias$^{3}$, Elena De Momi$^{2}$, Pietro Balatti$^{1}$, and Arash Ajoudani$^{1}$
\thanks{$^1$HRI$^2$ Lab, Istituto Italiano di Tecnologia, Genoa, Italy. {\tt\small hamidreza.raei@iit.it}}%
\thanks{$^{2}$ Department of Electronics, Information and Bioengineering, Politecnico di Milano, Milan, Italy.}
\thanks{$^{3}$ Robotics and Mechatronics lab, Systems Engineering and Automation Department, University of Malaga, Malaga, Spain.}%
\thanks{This work was supported by the European Commission's Marie Skłodowska-Curie Actions (MSCA) Project RAICAM (GA 101072634) and European Research Council's (ERC) starting grant Ergo-Lean (GA 850932).}
}

\markboth{Journal of \LaTeX\ Class Files,~Vol.~14, No.~8, August~2021}%
{Shell \MakeLowercase{\textit{et al.}}: A Sample Article Using IEEEtran.cls for IEEE Journals}



\maketitle

\begin{abstract}
This letter introduces an innovative visuo-haptic interface to control Mobile Collaborative Robots (MCR). Thanks to a passive detachable mechanism, the interface can be attached/detached from a robot, offering two control modes: local control (attached) and teleoperation (detached). These modes are integrated with a robot whole-body controller and presented in a unified close- and far-proximity control framework for MCR. The earlier introduction of the haptic component in this interface enabled users to execute intricate loco-manipulation tasks via admittance-type control, effectively decoupling task dynamics and enhancing human capabilities. In contrast, this ongoing work proposes a novel design that integrates a visual component. This design utilizes Visual-Inertial Odometry (VIO) for teleoperation, estimating the interface's pose through stereo cameras and an Inertial Measurement Unit (IMU). The estimated pose serves as the reference for the robot's end-effector in teleoperation mode. Hence, the interface offers complete flexibility and adaptability, enabling any user to operate an MCR seamlessly without needing expert knowledge. In this letter, we primarily focus on the new visual feature, and first present a performance evaluation of different VIO-based methods for teleoperation. Next, the interface's usability is analyzed in a home-care application and compared to an alternative designed by a commercial MoCap system. Results show comparable performance in terms of accuracy, completion time, and usability. Nevertheless, the proposed interface is low-cost, poses minimal wearability constraints, and can be used anywhere and anytime without needing external devices or additional equipment, offering a versatile and accessible solution for teleoperation.

\end{abstract}




\section{Introduction}


\PARstart{M}{obile} Collaborative Robots (MCR), often referred to as mobile cobots, are garnering substantial attention not limited to manufacturing and logistics but also expanding into the domains of healthcare, and in-home care~\cite{allaban-2020}. Such widespread interest stems from the remarkable manipulation and stable locomotion capabilities inherently embedded in these robotic systems. However, despite their potential, the large-scale adoption of MCRs has been hindered by their lack of autonomy. This has been a major barrier to their integration into complex and dynamic environments. In response, significant efforts are devoted to increasing the MCRs' decisional and interaction autonomy~\cite{engemann2020omnivil}.


\begin{figure}
\centering
    \includegraphics[trim=0.2cm 0.2cm 0.2cm 0.2cm, clip, width=0.5\textwidth, height=0.2\textheight]{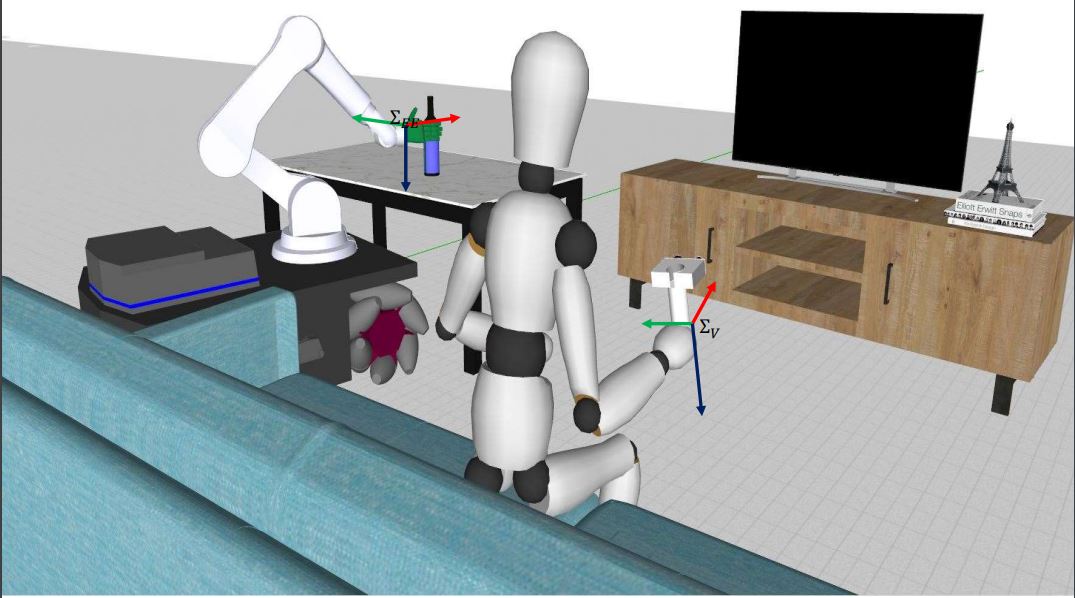}
    \caption{The proposed VIO-based teleoperation interface enables individuals to control MCRs in both close and distant settings, facilitating the performance of Activities of Daily Living (ADLs) in home care environments.}
    \label{fig:concept}
\end{figure}

In a parallel and synergistic effort, researchers are developing intuitive interfaces to enable human-in-the-loop control of MCRs, facilitating their use in close proximity and remote scenarios for effort-demanding or potentially hazardous tasks. In a step towards this goal, we have recently developed a collaborative framework that enables the MCR to function as a supernumerary mobile limb,    
leveraging its robust capabilities in tandem with human perceptual faculties~\cite{giammarino-2022}. Through a portable admittance-type interface and a priority-based whole-body controller, a user can move the arm, the mobile base, or a combination of both to perform complex loco-manipulation tasks~\cite{gandarias2022enhancing}. 

With the new objective of enabling users to control the robot's locomotion and manipulation from a remote location (near or far from the MCR), especially for tasks such as assisting senior adults with limited mobility in their daily activities, this study introduces a novel design for an MCR teleoperation interface (see Fig. \ref{fig:concept}). To eliminate the need for any additional settings and external equipment, we propose to include a set of stereo cameras, and an inertial measurement unit (IMU) in the device for implementing visual-inertial odometry (VIO) to infer human commands.  

In more details, this study proposes a novel visuo-haptic interface for MCR control, prioritizing critical factors essential for an efficient teleoperation interface, including portability, cost-effectiveness, independence from external sensors, and the absence of wearability requirements. In summary, the contributions of this work can be listed as follows:

\begin{itemize}
	\item A passive mechanism that allows attaching and detaching the interface from the MCR without requiring additional actuation systems. This mechanism enhances adaptability and allows users to switch seamlessly from teleoperation to local operation.
	\item The integration of a VIO method into a low-cost wireless device that can be used for teleoperation. The method has been integrated into simple, inexpensive, and easily accessible hardware. It uses a stereo camera system and an IMU to estimate the position of the interface without the need for external sensors or complex and expensive tracking systems. This solution allows users to perform teleoperation tasks in an infinite workspace without limiting their movement.

	\item A performance evaluation of different hardware and VIO approaches to assess teleoperation accuracy according to several metrics is also presented. 
	\item Integration of the VIO-based teleoperation approach into our previous haptic-based framework \cite{giammarino2023open}, both from hardware and software points of view. Thus, proposing a complete and unified framework that allows control both remotely, i.e., with the interface detached, thanks to the visual part of the interface, and locally, i.e., with the interface attached, thanks to the haptic part.
	\item Experimentation of the entire framework through a usability study and comparison with a state-of-the-art wireless teleoperation solution based on an IMU-based motion capture system.

        \item A detailed discussion on the advancements of our proposed solution incorporates both quantitative and qualitative metrics. These considerations are based on the results and performance revealed by the aforementioned experimental analyses.
\end{itemize}

The remainder of the manuscript is organized as follows: Section II includes state-of-the-art analysis of teleoperation frameworks. The proposed framework is presented in section III. Section IV depicts the experimental setup. Then, the results of the experiments are presented and discussed in depth. Finally, the conclusions are presented in section VI.

\section{Related Works}

Teleoperation technology has found diverse applications in robotics, spanning domains such as search and rescue~\cite{perez2017semi}, industrial inspection~\cite{rocha-2021}, surgery~\cite{leven2005telerobotic}, disaster relief~\cite{8594509}, and space exploration~\cite{bluethmann-2003}. 



Among all interfaces used for teleoperation, the joystick stands out as a prevalent teleoperation tool, extensively employed in various studies \cite{mavridis2015subjective, solanes2020teleoperation}. Despite its widespread adoption, teleoperating a system with a high-DoF using a low-DoF interface poses a substantial challenge, as highlighted in the work of Garate et al. (2021) \cite{garate2021scalable}. Furthermore, employing a joystick with a high-DoF could compromise the intuitiveness of the teleoperation interface \cite{dekker2023design}. An innovative approach to address this challenge involves embedding the robot's high-DoF dexterous behavior into low-DoF latent actions, facilitating human control, as illustrated in the study by Jeon et al. (2020) \cite{jeon-2020}. 
Despite its efficacy, this method induces task dependency and hinders adaptability to other applications.

Alternatively, Inertial and optical motion capture systems offer intuitive and accurate solutions for teleoperating various robots, ranging from robotic arms\cite{mavridis-2011}, and MCRs \cite{wu-2019}, to humanoid \cite{8464136} and legged robots\cite{zhou2022teleman}. However, It is important to point out, that the integration of motion capture systems as a teleoperation solution is usually expensive and may obstruct adaptability to diverse environments since it has wearability constraints and in the case of optical systems, necessitates the installation of external sensors within the operational environment.





Other solutions for teleoperation include employing single or multiple cameras to track specific markers~\cite{giammarino2023open} or to estimate the human-body pose in a markerless manner~\cite{hirschmanner2019virtual, fortini2023markerless}. Compared to motion capture systems, these alternatives offer more economical solutions at the cost of lower accuracy, whilst, their functionality remains dependent on the external sensors within the operational environment.


As previously mentioned, the primary objective of this work is to develop a non-wearable, portable, cost-effective, and user-friendly interface that does not rely on environmental sensors or external markers. Unlike previous studies, our approach aims to overcome the aforementioned limitations and broaden the applicability of the teleoperation interface to control MCRs when attached, augmenting the capabilities of our previous haptic interface~\cite{gandarias2022enhancing}.

\begin{figure*}[!htb]
    \centering
    \includegraphics[trim=0 3.2cm 0 3.5cm, clip, width=0.95\textwidth, height=0.35\textheight]{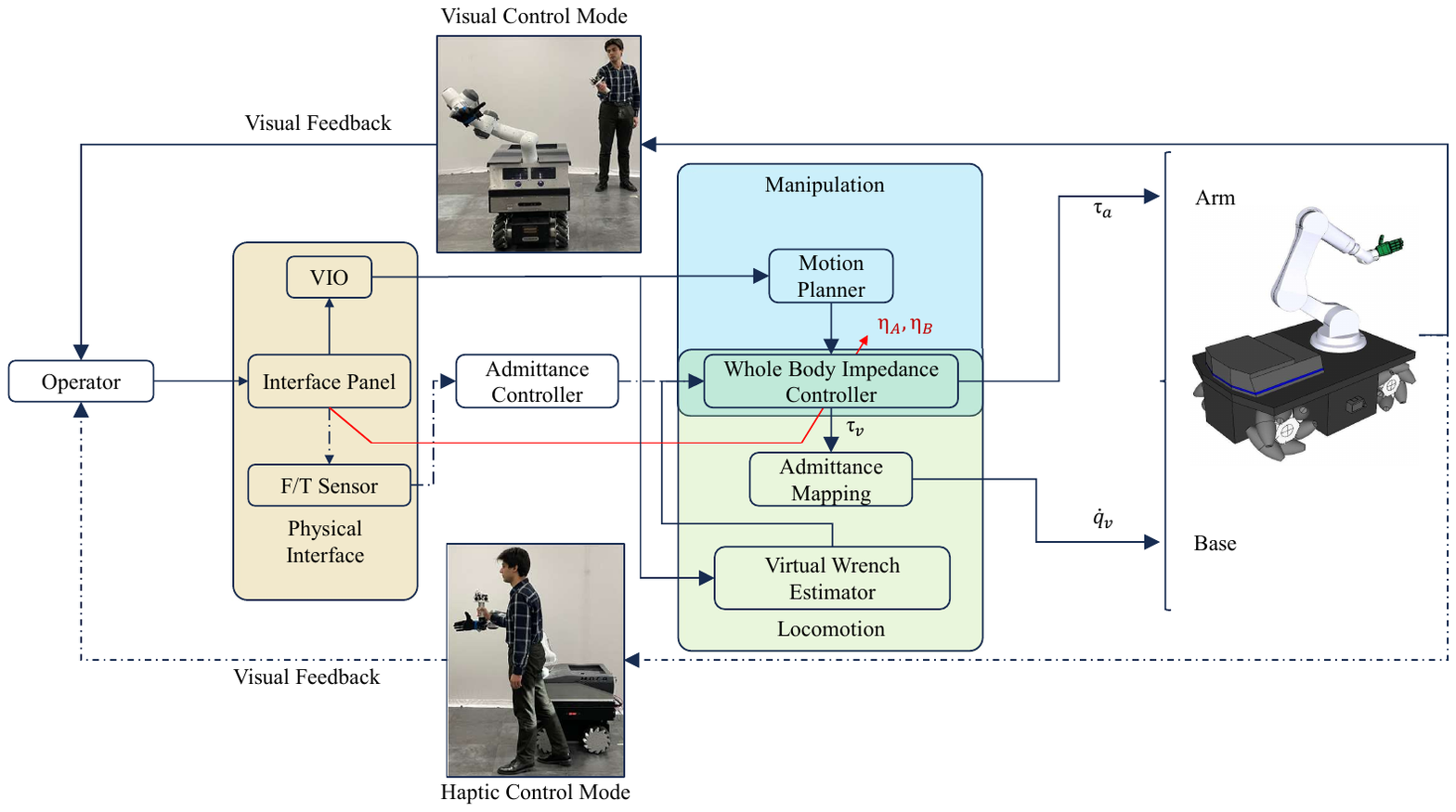}
    \caption{Block diagram of the extended visuo-haptic framework, to be used for MCRs under a whole-body impedance controller. Dashed lines represent connections related to the haptic part whilst solid lines represent those related to the visual part.}

    \label{fig:block_diagram}
\end{figure*}

\section{Framework Overview}

\subsection{\textbf{The Open Multi-Purpose Interface (OMI)}}
The proposed OMI serves as both a haptic and visual interface, allowing remote or local robot operation, as depicted in the system overall block diagram in Fig. \ref{fig:block_diagram}. It facilitates local operation via a haptic interface when attached to the end-effector. When detached, it employs the visual interface for teleoperation, details of which are explicitly outlined in the following subsections.

\subsubsection{\textbf{Haptic Interface}}
This interface is built around a System on a Chip (SoC) solution, specifically the M5Core2 which utilizes an ESP-32 microcontroller and incorporated a touchscreen, and an F/T sensor. It excels in user-friendliness, enabling one-handed local operation that makes the OMI flexible and programmable. Users can efficiently oversee a range of robot functions and easily adapt to specific task needs. Notable features of this haptic interface include the capability to switch motion modes, manage the gripper, adjust task motion priorities between the arm and the base, or set the impedance. Specific details on the use of the haptic interface and its integration with the whole body controller are illustrated in our previous studies~\cite{gandarias2022enhancing,giammarino-2022,giammarino2023open}.


\subsubsection{\textbf{Visual Interface}}
Leveraging the VIO output, the visual interface interprets the motion of the interface to formulate a reference command for the robot. This capability is embedded in an all-in-one handheld device that is well-suited for teleoperation in any environment. The main idea is to perform VIO on the handheld device, the OMI, and eliminate the need for an external sensor in the environment to localize the interface with regard to its initial pose.

Several algorithms can be used to perform VIO, including RTAB-MAP~\cite{labbe2019rtab}, LSD-SLAM~\cite{engel2014lsd}, and ORB-SLAM~\cite{mur2017orb}, which are typically compatible with RGB-D and stereo cameras. However, one of the specific differences of the RTAB-MAP algorithm is its reliance on Appearance-Based loop-closure, as opposed to the dependence on geometric features or landmarks, used mostly in other algorithms.
This special feature enhances the performance of the algorithm in environments with visual patterns but not many clear geometric landmarks. Implementing a technique known as the ``bag of words'', significantly improves loop closure detection while remaining efficient in terms of memory usage and computational load. Thereby mitigating drift by promptly recognizing previously visited scenes\cite{labbe2019rtab}.

RGB-D and stereo cameras stand out as predominant choices for providing visual, and geometric data to VIO, owing to their widespread availability and compatibility with a majority of VIO algorithms. The type of visual sensors and the communication between the interface and the robot impact the accuracy and frame rate of the VIO system, influencing computational demands and data transmission requirements. Therefore, empirical testing is crucial to validate real-time teleoperation and precision of various setups.

\subsection{\textbf{Robot Whole-body Cartesian Impedance Controller}}
According to the formulation presented in~\cite{kim2020moca}, the MCR can perform two distinct behaviors: manipulation and locomotion. In the context of the Manipulation mode, when the mobile platform is deemed to be in a quasi-static state and a virtual inertial  and damping representation is adopted for the entire robot, effectively replacing the inertial and damping parameters of both the arm and the mobile platform with their virtual counterparts, the resulting equation of overall dynamic of the MCR can be expressed as follows:
\begin{equation}
    M_{adm} \Ddot{q} + D_{adm} \dot{q} = \tau_{c} + \tau_{ext}
\end{equation}
which can be expanded as below:

\vspace{5 pt}
\textit{$
\begin{pmatrix}
    M_{v} & 0 \\
    0 & M_r(q_r)
\end{pmatrix}
\Ddot{q}
+
\begin{pmatrix}
    D_{v} & 0 \\
    0 & C_r(q_r, \dot{q_{r}})
\end{pmatrix}
\dot{q}
+
\begin{pmatrix}
    0 \\
    g_r
\end{pmatrix}
= $}
\vspace{1 mm}
\begin{equation}
\tau_{c}
+
\tau_{ext}
\end{equation}


Here, \textit{$g_r$} is the gravity vector, \textit{$M_{adm} \in \mathbb{R}^{m \times m} $} and \textit{$D_{adm} \in \mathbb{R}^{m \times m} $} are the virtual inertial and virtual damping, and \textit{$q = (q_v, q_r)^T \in \mathbb{R}^{10} $} represents whole-body joint variables, and \textit{$\tau_{c} = [ \tau_{v}^{T}    \    \tau_{r}^{T}]^{T} $} and \textit{$\tau_{ext} = [0^{T} \ \tau_{r, ext}^{T}]^{T} $} are respectively the high-level commanded torque vector 
and the external torque vector that is passed for the mobile admittance controller (\textit{$ \tau_v $}) and the arm low-level controller (\textit{$ \tau_{r} $}). 


The whole-body controller's advantage lies in its ability to execute operator-commanded motions by prioritizing specific joint movements. In other words, a weighting matrix is established to compute the high-level torque references(\textit{$\tau_c$}) as explained in our previous study\cite{kim2020moca}. This weighting matrix is determined as follows:
\begin{equation}
    W(q) = H^{T} M^{-1}(q)  H
\end{equation}

Here, the selection of \textit{$H$} is based on the specific task at hand, whether it involves locomotion or manipulation, and it is constructed as a diagonal matrix:

\begin{equation}
    H = 
    \begin{pmatrix}
        \eta_{_B} I_{n_b} & 0_{n_{b}\times n_{a}} \\
        0_{n_{b}\times n_{a}} & \eta_{_A} I_{n_a}
    \end{pmatrix}
\end{equation}
where \textit{$\eta_{_B}$} and \textit{$\eta_{_A}$}  penalize the motion of base or arm, respectively. These parameters can be adjusted either dynamically or manually to prioritize the movement of either the arm or the base, depending on the specific task at hand. Nonetheless, for the haptic interface, the study mentioned earlier\cite{giammarino-2022} has clearly demonstrated its adaptability to both manipulation and locomotion. In the following, the adaptability of the visual interface to these modes is illustrated.


\subsubsection{\textbf{Manipulation using VIO}}

During manipulation, the priority of the motion is given to the robotic arm and it works under the whole-body controller. Our objective is to achieve the same motion for the arm end-effector w.r.t its initial pose, as motion in the teleoperation interface w.r.t the interface initial pose. The world coordinate system is denoted as $\Sigma_{W}$ and the transformation between the initial and the final pose of the end-effector is denoted as \textit{T}$^{\textit{EE}_f}_{\textit{EE}_i} $, whilst, the transformation between the initial and the final pose of the teleoperation interface is denoted as \textit{T}$^{\textit{V}_f}_{\textit{V}_i}$. Thus, our objective of the manipulation using VIO is to satisfy \textit{T}$^{\textit{EE}_f}_{\textit{EE}_i} $ = \textit{T}$^{\textit{V}_f}_{\textit{V}_i}$. 



The importance of a whole-body controller is emphasized when the operator extends the arm beyond its typical range. In such cases, the controller commands a coordinated motion involving both the manipulator and the mobile platform, ensuring that the end-effector attains the desired pose.

The mobile platform's odometry provides the transformation from the mobile platform's initial pose to its pose at any given moment,
denoted as \textit{T}$^{\textit{M}}_{\textit{W}}$. The \textit{${T^{R}_{M}}$} refers to the transformation from the mobile platform coordinate system to the Arm's base. Also, \textit{$q = (q_v, q_r)^T \in \mathbb{R}^{10} $} represents whole-body joint variables, including 7 DoF for the arm shown as \textit{$q_r$} and 3 DoF of the mobile platform shown as \textit{$q_v = (q_{vx}, q_{vy}, q_{vz})$}. The command to the MCR is the transformation from $\Sigma_{W}$ to the final pose of the end-effector, which is supposed to replicate the motion from the interface while scaling the translational motion to accommodate the specific needs of the application. Hence, the transformation is obtained as follows:
\begin{equation}
\begin{aligned}
{T^{EE_f}_{W}}(q) &= {T^{M}_{W}}(q_v) \ {T^{R}_{M}} \ {T^{EE_i}_{R}}(q_r) \ {T^{EE_f}_{EE_i}}(q_r) 
\end{aligned}
\end{equation}


The scaling factor governing translation at the end-effector is dictated by the parameter denoted as $\alpha$. This parameter is applied through element-wise multiplication, as expressed by the following:

\begin{equation}
\begin{aligned}
{T^{EE_f}_{EE_i}} &=  \begin{pmatrix}
1 & 1 & 1 & \alpha \\
1 & 1 & 1 & \alpha \\
1 & 1 & 1 & \alpha \\
1 & 1 & 1 & 1 \\
\end{pmatrix} \odot {T^{v_f}_{v_i}} 
\end{aligned}
\end{equation}



\subsubsection{\textbf{Locomotion using VIO}}

During locomotion, the motion priority shifts to the mobile platform in the whole-body controller. The displacement of the teleoperation interface turns into a virtual force and Torque at the robot's end effector respectively denoted as \textit{$F_{vir} = \begin{pmatrix}
    F_x \ F_y \ F_z
\end{pmatrix}^{T}$} and \textit{$\tau_{vir} = \begin{pmatrix}
    \tau_x \ \tau_y \ \tau_z
\end{pmatrix}^{T}$}, passing through the admittance controller, resulting in the mobile platform motion. Force and torque at the end effector are virtually obtained w.r.t to $\Sigma_{W}$ coordinate system. In the following equations, linear stiffness and rotational stiffness are denoted as K and C respectively.
\begin{equation}
    \begin{pmatrix}
        F_x \\
        F_y \\
        F_z \\
    \end{pmatrix} = \begin{pmatrix}
        K_x & 0 & 0 \\ 
        0 & K_y & 0\\
        0 & 0 & K_z\\ 
    \end{pmatrix} \ \begin{pmatrix}
        \Delta x\\
        \Delta y\\
        \Delta z\\
    \end{pmatrix}
\end{equation}
\begin{equation}
    \begin{pmatrix}
        \tau_x \\
        \tau_y \\
        \tau_z \\
    \end{pmatrix} = \begin{pmatrix}
        C_x & 0 & 0 \\ 
        0 & C_y & 0\\
        0 & 0 & C_z\\ 
    \end{pmatrix} \ \begin{pmatrix}
        \Delta \phi \\
        \Delta \theta \\
        \Delta \psi \\
    \end{pmatrix}
\end{equation}


In locomotion, the mobile platform moves on the surface, involving 3 DoF which results in (\textit{$K_z = C_x = C_z = 0$}). A circular virtual threshold is set for locomotion, requiring operators to cross it initially to map a virtual wrench onto the robot's end-effector. Additionally, a maximum radius is defined, denoted as \textit{$R_m$} to limit the magnitude mapped wrench if the operator surpasses safety constraints \textit{$F_{vir} \in {[-K R_m, K R_m]} $}. 



We observed that attempting diagonal movements with the robot caused several subjects to encounter various challenges, resulting in a modification that mandates robot motion in only one direction at a time, and changes in direction are only allowed after it comes to a stop. To optimize accuracy, the selected direction aligns with the predominant interface's motion axis, commanded by the operator. 


This algorithm is implemented to enable long-distance locomotion without requiring the operator to physically move comparable to the mobile platform. Once the virtual wrench in this process is replaced by the output of the F/T sensor installed on the end-effector, the locomotion part is similar to the haptic interface\cite{kim2020moca}. The control block diagram of the teleoperation interface for both locomotion and manipulation mode is illustrated in Fig. \ref{fig:block_diagram}.




\section{The Experimental Setup}

This section outlines two types of experiments designed to assess the effectiveness of the VIO interface. In the first experiment, we compare the proposed VIO interface against alternative VIO configurations, evaluating their respective accuracies and update rates. The second experiment involves a comparative analysis of the proposed VIO interface against a well-established teleoperation method employed for mobile robots, in the application of home-care scenarios.





\subsection{\textbf{VIO Comparison}} 

This comparison was made to compare various setups for VIO in terms of their accuracy, given a specific motion. The experiment involves translational and rotational motions along/around the x, y, and z axes, aimed to compare the performance of two VIO based setups. A structure was designed to link the VIO setups to the Franka Emika robot end-effector. This link facilitates precise motion control in different directions and
orientations and enables accurate comparisons of various
setups and configurations by providing ground truth data of the position and the orientation of the robot end-effector.

The wired stereo setup is a ZED2i, stereo camera, that streams images at 30 Frame Per Second (FPS) with a built-in IMU and a magnetometer. This camera features a Field of View (FoV) of 72 degrees horizontally and 44 degrees vertically. The RGB-D wired setup is an Intel Depth Camera D435i, it streams images at 30 FPS, and the FoV of 69 degrees horizontally and 42 degrees vertically.

For the wireless configuration of the RGB-D setup, the Intel D435i is connected to a Raspberry Pi 4, which streams images and depth frames via WiFi. The main difference between the wired RGB-D setup and the wireless one lies in its limited frame rate capability, with the wireless configuration achieving a maximum streaming rate of 15 FPS.

\begin{figure}[!htb]
    \centering
    \includegraphics[width=0.45\textwidth, height=0.25\textheight]{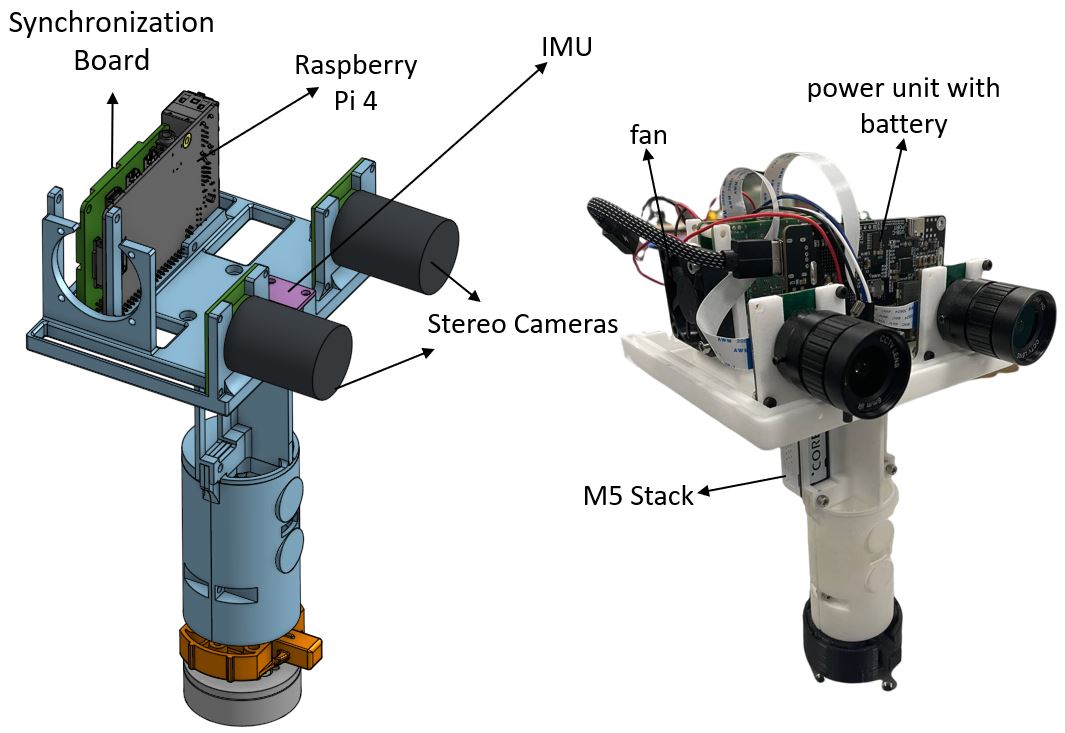}
    \caption{The image showcases the designed model of the interface on the left, with the corresponding assembled version of the interface presented on the right.}
    \label{fig:design_interface}
\end{figure}

The wireless stereo setup is designed, by prioritizing FPS, weight, and the system latency, in addition to constraints on power consumption and computational resources. The proposed wireless VIO interface consists of a Raspberry Pi 4, an Arducam synchronized stereo kit (two 12.3MP 477P camera modules), an M5Core2, an IMU (MPU-9250), and a power unit with batteries. The proposed stereo interface has a FOV of 65 degrees horizontally, and 40 degrees vertically and could stream up to 30 FPS. Additionally, the CAD (Computer-Aided Design)
representation of the interface is displayed, along with the
assembled interface depicted in Fig. \ref{fig:design_interface}. The CAD design is accessible through the following link: \linebreak 
\href{https://cad.onshape.com/documents/a6ae72c5d9abcc1a94e60661/w/57a68a004590754820b853d7/e/d85df546bfce8615c5f0752f?renderMode=0&uiState=65ad295fab972d5c4fa47903}{\texttt{https://cad.onshape.com/documents/a6ae72c}}

Additionally, access to this study source code is available on our github using the following link: 

\url{https://github.com/hrii-iit/hrii_vo.git}

\subsection{\textbf{Home-Care Application}}
This section outlines the scenario chosen to validate the proposed teleoperation interface against a widely used motion capture system, employed in diverse studies for teleoperating various robots such as mobile collaborative robots\cite{wu-2019} and legged robots\cite{zhou2022teleman}. The motion capture system employed in this study is the Xsens system, which incorporates multiple IMUs affixed to wearable costumes. Several subjects were assigned to perform this scenario using teleoperation with both methods.

\subsubsection{\textbf{Experiment Scenario}}

The determined scenario for this experiment entails the subject executing three sequential tasks as follows: 


\begin{enumerate}[i.]
    \item \textit{{Grasping a relatively small ball in manipulation mode.}}
    \item \textit{{Locomoting the mobile platform from its initial position to the vicinity of the drawer.}}
    \item \textit{{Depositing the ball into the drawer and subsequently closing the drawer.}}
\end{enumerate}

\begin{figure}[!htb] 
    \centering
    \includegraphics[trim=0 5.0cm 0 4.5cm, clip, width=0.45\textwidth, height=0.18\textheight]{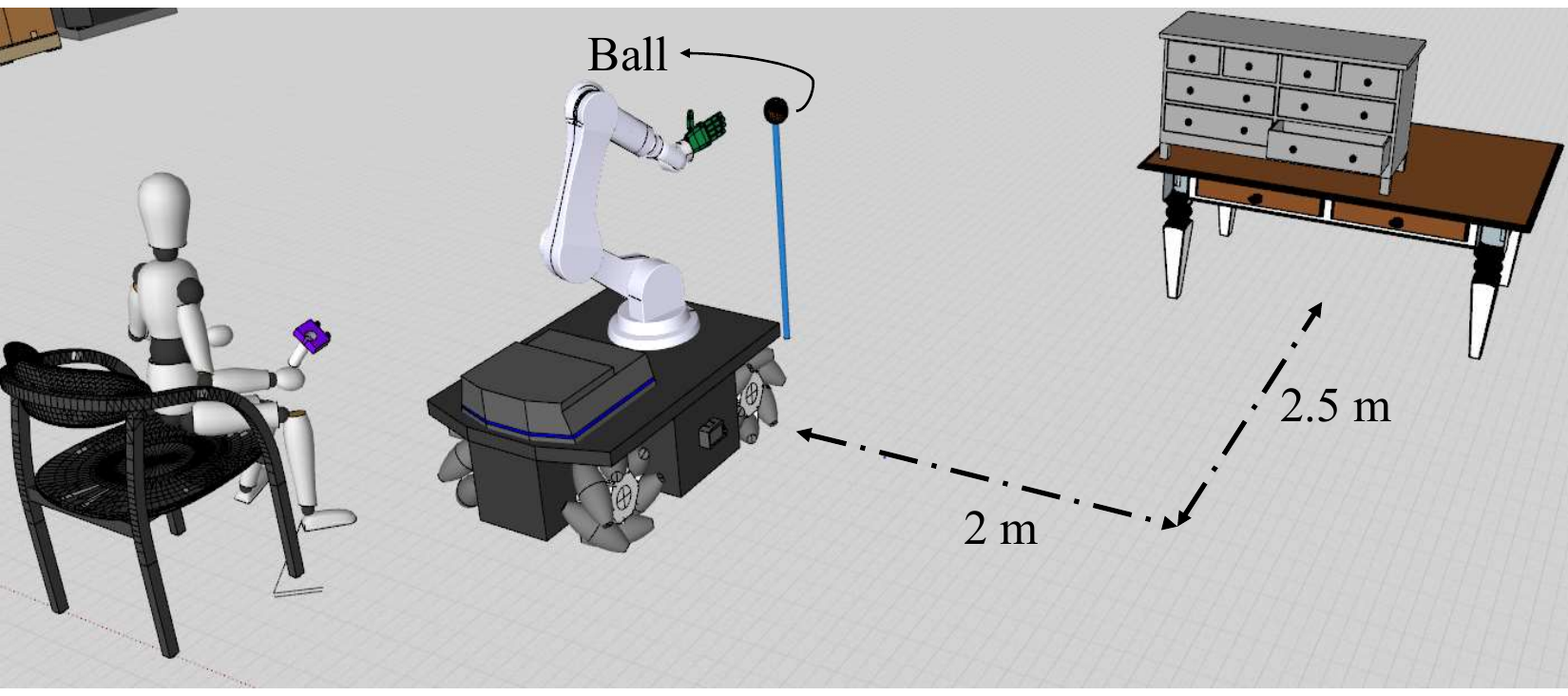}
    \caption{A sketch of the experimental setup, used for home-care scenario application.}
    \label{fig:exp_setup}
\end{figure}

In Fig. \ref{fig:exp_setup} the experiment setup and distances between the starting point and the drawer are illustrated.

\subsubsection{\textbf{Study Protocol}}

During the comparative study, subjects were asked to sit on a chair and perform remotely while looking at the robot on their side. Participants were asked to perform the sequential tasks two times, unaware of the interface type assigned for each trial, and they wore the Xsens suit and grasped the VIO interface throughout both trials. The allocation of each interface was randomized for each participant. This study sought to evaluate these teleoperation systems based on both quantitative and qualitative parameters. The specified quantitative parameters included in this study are listed as follows:

\begin{itemize}
    \item {The success rate, which is determined by dividing the number of successfully completed subtasks by the total trials across all subtasks for each participant}.
    \item {The corresponding time duration (${T_c}$) of overall performance for each subject which is the sum of the time duration for all three successful subtasks.}
    \item {Average preparation time for each subject to use the teleoperation system denoted as ${T_p}$}.
\end{itemize}

Additionally, to assess the effectiveness of the proposed teleoperation interface in terms of qualitative parameters, the NASA Task Load Index (TLX) questionnaire scores were obtained from each subject upon completion of all three sequential tasks using each teleoperation system.



\section{Results and Discussion}


\begin{figure*}[!htb]
    \centering
    \includegraphics[trim=0 7.0cm 0 6.0cm, clip, width=1.0\textwidth, height=0.25\textheight]{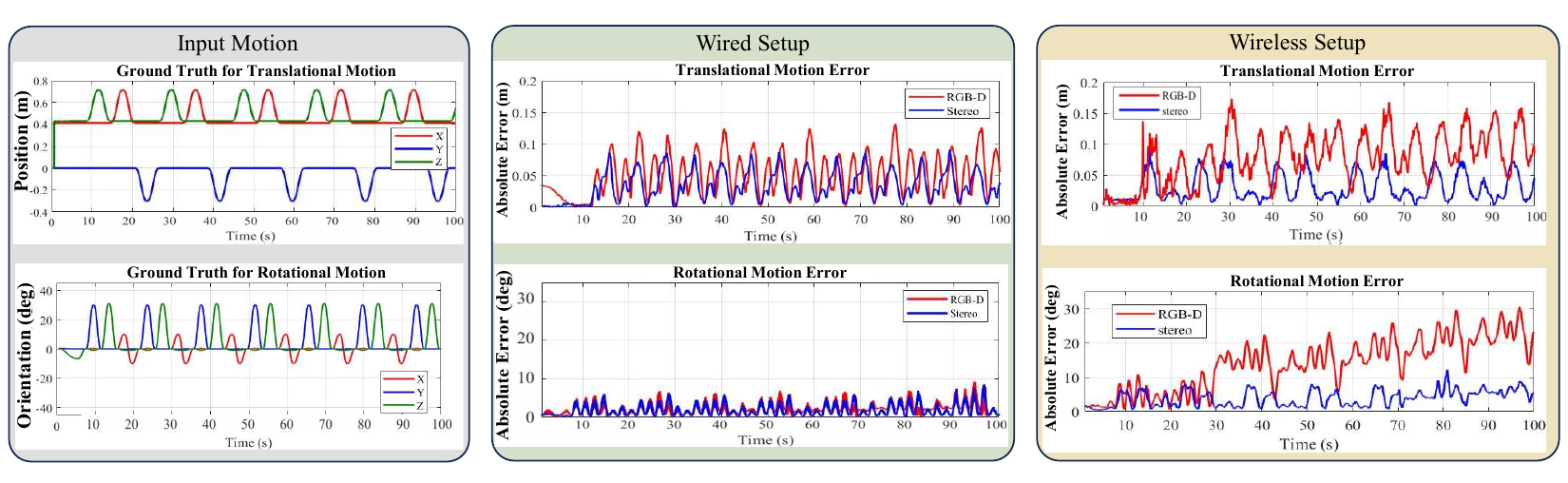}
    \caption{This figure showcases the VIO performance using two sensors, Stereo and RGB-D, in two setups: wired (left column labeled as (a)) and wireless (right column labeled as (b)). The bottom plots in both columns display the absolute error in position estimation, with the red line representing the RGB-D and the blue line representing the stereo setup.}
    \label{fig:VO_compare_res}
\end{figure*}

\subsection{\textbf{VIO Comparison}}

The translational and orientational motion given to the VIO setup attached to the end-effector has been plotted in Fig. \ref{fig:VO_compare_res} as input motion, which is considered as ground truth since it is measured through inverse kinematics of the robotic arm.

All mentioned setups in wired and wireless configurations were tested with the same motions and the VIO estimation of the motion is compared to the ground truth. Thus, the absolute error of the RGB-D and the stereo setup have been plotted for both wireless and wired configurations in Fig. \ref{fig:VO_compare_res}. 

 For the case of wired configurations, although in translational motion higher error for the RGB-D setup has been observed, the orientational motion does not exhibit a considerable difference between stereo and RGB-D setup.

In the wireless configuration, notable errors are apparent in the RGB-D camera, surpassing those observed in the stereo setup. These errors escalate over time, signaling a drift phenomenon affecting both translational and orientational motions, with a more pronounced impact on orientational accuracy. In this experimental context, while the difference in translational motion, between the wired and wireless setups is significant for the RGB-D camera, the stereo camera exhibits relatively similar errors in both configurations. The average error during the experiment for the RGB-D in wired and wireless setups are 0.0541 and 0.0833, and for the stereo setups are 0.0365 and 0.0384 for the wired and wireless configurations respectively.

It is worth mentioning that regarding the update rate of the VIO output achieved by the teleoperation interface, for the wired configuration, the odometry update rate is consistent at approximately 30 Hz for both setups. However, wireless setups face several challenges ranging from computation limitations to bandwidth of wireless communication. Considering a trade-off between the update rate and the image resolution resulting in the accuracy of the odometry, we could reach 30 Hz for the stereo setup and 15 Hz for the RGB-D camera.

\begin{figure}[!htb] 
    \centering
    \includegraphics[trim=5.5cm 3.8cm 6.5cm 3.8cm, clip, width=0.48\textwidth, height=0.26\textheight]{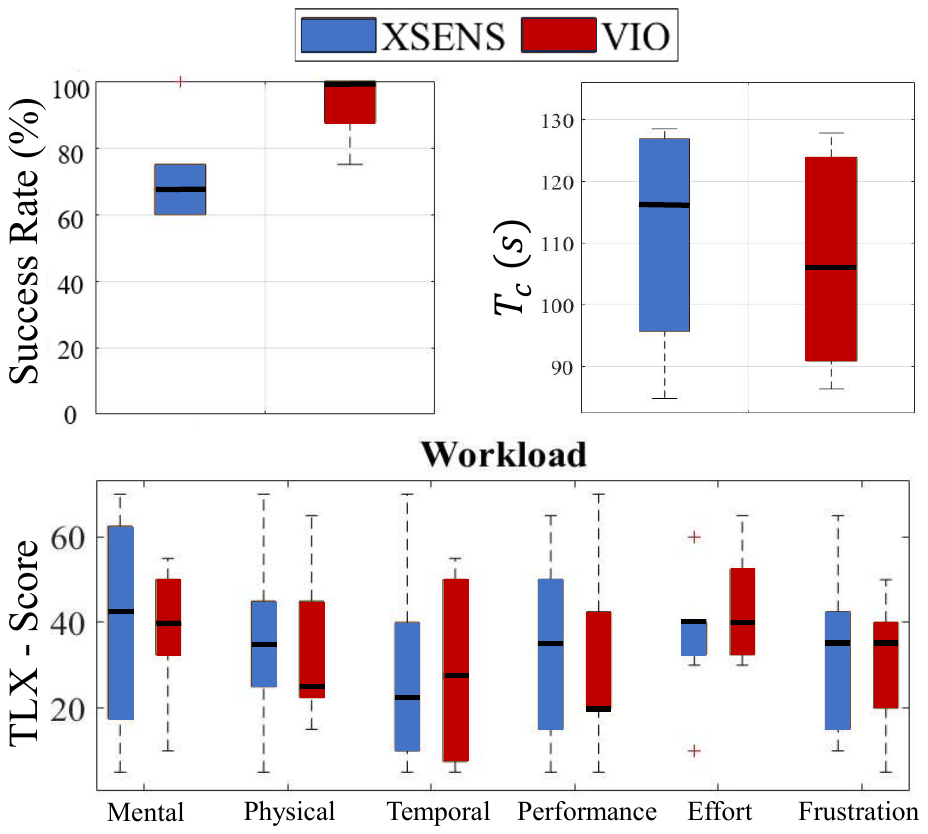}
    \caption{Illustrates the comparison between qualitative and quantitative parameters of the proposed VIO interface and Xsens system performance, used for teleoperation in the home-care application scenario}

    \label{fig:NASA}
\end{figure}

\begin{figure*}[!htb]
    \centering
    \includegraphics[trim=0 2.5cm 0 2.5cm, clip, width=1.0\textwidth, height=0.415\textheight]{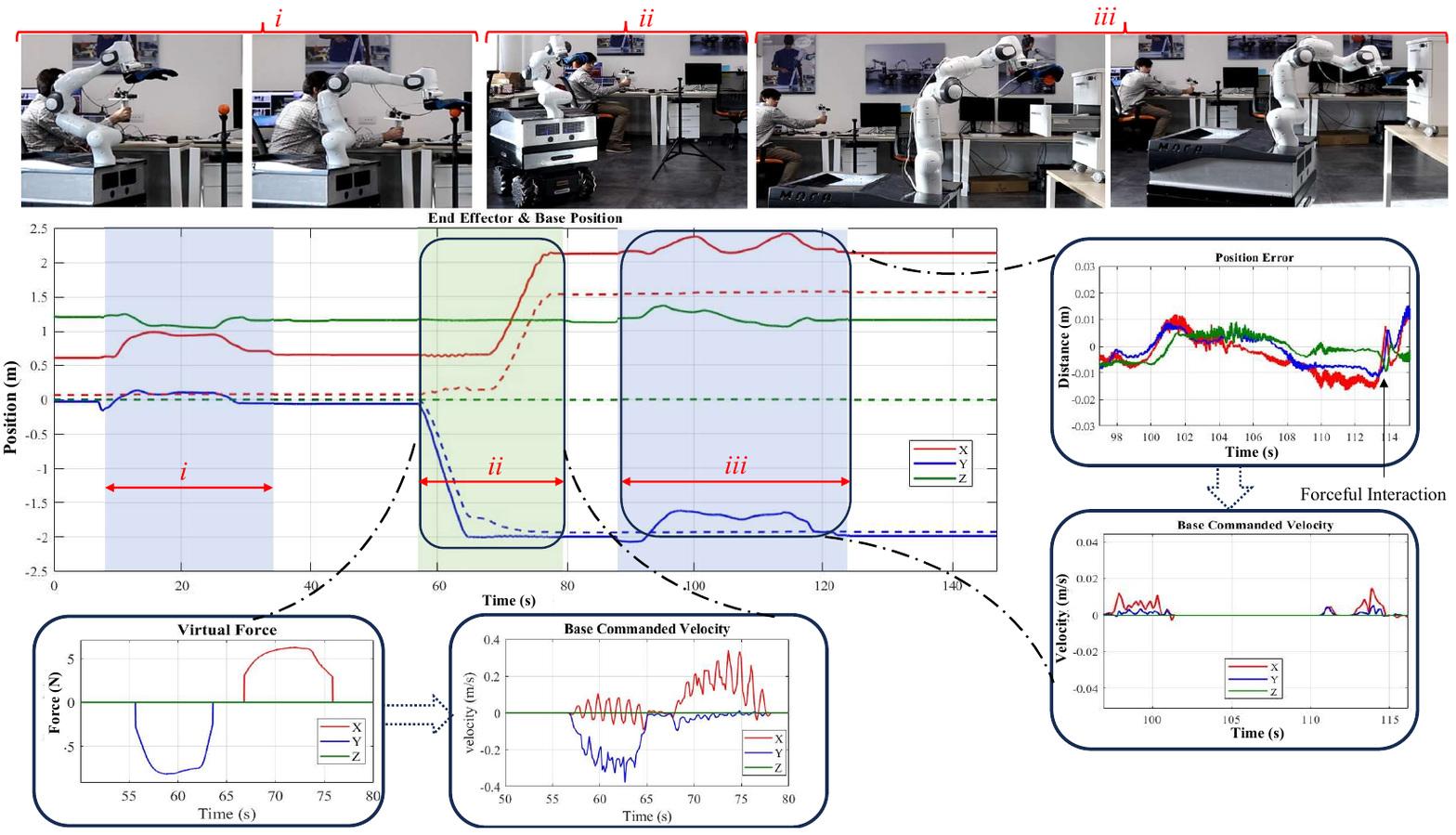}
    \caption{The figure illustrates VIO interface performance in the Home-Care Scenario, with dashed lines for the mobile platform position and solid lines for the End-Effector Position. The blue highlight indicates manipulation mode, and the green one signifies locomotion mode.}
    \label{fig:Demo_plot}
\end{figure*}

\subsection{\textbf{Home-Care Application}}

Eight participants performed the Home-Care application scenario utilizing the proposed VIO interface and the Xsens system. In Fig. \ref{fig:NASA}, the depicted success rates for each interface highlight a higher success rate with the VIO interface compared to the Xsens system in the designated scenario. The average success rate for the Xsens system is 70.625\% and the average for the proposed VIO interface is 93.75\%. 


Regarding the corresponding time duration (${T_c}$) for each teleoperation system, both systems had similar time lengths. However, the median ${T_c}$ for the proposed VIO interface is slightly lower (106 seconds) compared to the Xsens system's median ${T_c}$ of 116 seconds, indicating a marginally more efficient task completion with the VIO interface which is depicted in Fig. \ref{fig:NASA} at the top right side of the figure. Regarding the teleoperation system's preparation time (${T_p}$), the Xsens system necessitates about 3 minutes for donning, plus an additional approximately 1 minute for calibration, a requisite for every inertial motion capture system. Conversely, the VIO interface, being handheld and not wearable, demands no calibration, enabling the subject to initiate usage within 30 seconds.

The NASA-TLX questionnaire result is depicted at the bottom of Fig. \ref{fig:NASA}, and a statistical analysis of TLX parameters is performed. A Wilcoxon signed-rank test was conducted on NASA-TLX data, employing a significance level of 0.05. Findings indicate no statistical significance in all NASA-TLX parameters between the two teleoperation systems under consideration. The p-values for the Wilcoxon signed-rank test are as follows: mental demand (0.8984), physical demand (0.5), temporal demand (0.9062), performance (0.6719), effort (0.3281), and frustration (0.8906). These results suggest comparable task load indices for both teleoperation systems.

While both systems exhibit reasonable and relative accuracy, we claim that providing subjects with awareness of the specific points used for manipulation or locomotion in the teleoperation paradigm enhances task performance. In contrast, the Xsens system lacks clarity regarding the designated segment on the subject's arm for mimicking motions using teleoperation, potentially leading to errors as subjects may not be aware of the precise manipulation points. That could possibly be the reason for a slightly higher success rate, in the VIO teleoperation interface.

The comparative analysis of the VIO teleoperation system and the Xsens system reveals their similar performance across quantitative and qualitative parameters. Notably, the VIO interface exhibits slightly superior results in terms of success rate, in the aforementioned home-care scenario. The VIO interface being relatively cost-effective in addition to the extended preparation time, calibration requirements, and the wearability constraint of the Xsens system would suggest that the VIO interface could be more adaptable to various environments and applications.

An example of performing this home-care application scenario has been performed and visualized in Fig. \ref{fig:Demo_plot}. In this figure, the motion of the end effector is depicted with solid lines, and the motion of mobile platform is depicted with dashed lines. Additionally, segments, where the operator executed tasks in manipulation mode, are highlighted in blue, while locomotion mode is represented in green. Throughout the locomotion mode, the virtual force acting on the end effector is graphed at the bottom left of Fig. \ref{fig:Demo_plot}, influencing the commanded velocity for the mobile platform depicted on the right side of the virtual force plot.

Another parameter that influences the commanded velocity for the mobile platform is the extension of the arm beyond its regular range, leading to an error between the commanded and actual arm position, which results in the motion of the mobile platform under the whole-body impedance controller. The integration of an impedance controller offers several advantages, such as demonstrating compliance—allowing slight yielding or deformation in response to external forces. On the right side of Fig. \ref{fig:Demo_plot}, the position error resulting from arm extension and forceful interactions is indicated. Where exerting force is required to close the drawer, the error between the commanded and actual position of the arm increases which leads to small motion in the mobile platform evident in the commanded velocity depicted on the bottom right side of Fig. \ref{fig:Demo_plot}.

The video showcasing the performance of the proposed VIO interface is available at the following link: \href{https://youtu.be/cZpI6Cp2Atc}{\texttt{https://youtu.be/cZpI6Cp2Atc}}.

\section{Conclusions}


This study introduced a novel and low-cost visuo-haptic interface for MCR featuring a passive detachable mechanism, allowing local loco-manipulation when attached and teleoperation when detached. The control framework included a whole-body impedance controller to enhance the interaction of the robot with external objects in the environment.

Additionally, a home-care scenario was determined to validate and compare the performance of the proposed VIO interface with one of the widely known means of teleoperation. Results showed that both systems performed similarly in terms of qualitative parameters concerning the human workload whilst, in terms of quantitative parameters, the VIO interface slightly outperformed the Xsens system. The evaluation result obtained in this experiment indicates that the primary objective of this study, which is the development of a teleoperation interface that operates without dependency on environmental sensors while ensuring precision, and wireless connectivity, has been achieved. It is worth mentioning that the proposed teleoperation interface is not limited to being used for mobile cobots only and can be seamlessly integrated with various other robots.

In the future, considering the accuracy of the proposed visual interface validated in this study, it can be integrated with learning from demonstrations (LfD) applications. Additionally, this system can be integrated with shared autonomy, through perception algorithms that can detect and identify objects in the close vicinity of the robot, and potentially be used to predict the operator's intention of a given command to reduce the workload on the human.

\balance

\bibliographystyle{IEEEtran}

\bibliography{main.bib}


 




\vfill

\end{document}